\documentclass{article}

\usepackage[square,comma]{natbib}
\setcitestyle{numbers,square}

     \usepackage[preprint]{neurips_2023}

\usepackage[utf8]{inputenc} % allow utf-8 input
\usepackage[T1]{fontenc}    % use 8-bit T1 fonts
\usepackage{hyperref}       % hyperlinks
\usepackage{url}            % simple URL typesetting
\usepackage{booktabs}       % professional-quality tables
\usepackage{amsfonts}       % blackboard math symbols
\usepackage{nicefrac}       % compact symbols for 1/2, etc.
\usepackage{microtype}      % microtypography
\usepackage{xcolor}         % colors
\usepackage{graphicx}
\usepackage{listings}
\usepackage{graphbox}
\usepackage{makecell}
\usepackage{amsmath}
\usepackage{comment}
\usepackage{array}

\usepackage{bm}
\usepackage[ruled,vlined]{algorithm2e}
\usepackage{hyperref}
\hypersetup{
	colorlinks=true,
        breaklinks=true
}

\usepackage{marvosym}
\makeatletter
\def\@fnsymbol#1{\ensuremath{\ifcase#1\or \dagger\or \ddagger\or
   \mathsection\or \mathparagraph\or \|\or **\or \dagger\dagger
   \or \ddagger\ddagger \else\@ctrerr\fi}}
\makeatother

\newcommand{\mcbed}{\includegraphics[scale=0.4,align=c]{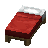}}
\newcommand{\mcbeef}{\includegraphics[scale=0.4,align=c]{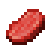}}
\newcommand{\mcbowl}{\includegraphics[scale=0.4,align=c]{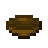}}
\newcommand{\mcbucket}{\includegraphics[scale=0.4,align=c]{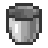}}
\newcommand{\mccarpet}{\includegraphics[scale=0.4,align=c]{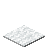}}
\newcommand{\mcchest}{\includegraphics[scale=0.4,align=c]{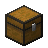}}

\newcommand{\mccobblestone}{\includegraphics[scale=0.4,align=c]{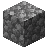}}
\newcommand{\mccobblestonewall}{\includegraphics[scale=0.4,align=c]{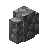}}
\newcommand{\mccookedbeef}{\includegraphics[scale=0.4,align=c]{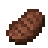}}
\newcommand{\mccookedmutton}{\includegraphics[scale=0.4,align=c]{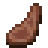}}
\newcommand{\mccow}{\includegraphics[scale=0.4,align=c]{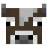}}
\newcommand{\mccraftingtable}{\includegraphics[scale=0.4,align=c]{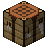}}
\newcommand{\mcdiamondsword}{\includegraphics[scale=0.4,align=c]{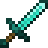}}
\newcommand{\mcfurnace}{\includegraphics[scale=0.4,align=c]{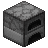}}
\newcommand{\mcironingot}{\includegraphics[scale=0.4,align=c]{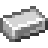}}
\newcommand{\mcitemframe}{\includegraphics[scale=0.4,align=c]{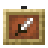}}
\newcommand{\mclever}{\includegraphics[scale=0.4,align=c]{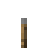}}
\newcommand{\mclog}{\includegraphics[scale=0.4,align=c]{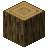}}
\newcommand{\mcmilkbucket}{\includegraphics[scale=0.4,align=c]{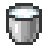}}
\newcommand{\mcmutton}{\includegraphics[scale=0.4,align=c]{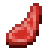}}
\newcommand{\mcpainting}{\includegraphics[scale=0.4,align=c]{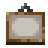}}

\newcommand{\mcshears}{\includegraphics[scale=0.4,align=c]{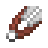}}
\newcommand{\mcsheep}{\includegraphics[scale=0.4,align=c]{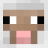}}
\newcommand{\mcsign}{\includegraphics[scale=0.4,align=c]{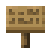}}
\newcommand{\mcstick}{\includegraphics[scale=0.4,align=c]{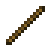}}
\newcommand{\mcstonepickaxe}{\includegraphics[scale=0.4,align=c]{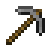}}
\newcommand{\mcstonestairs}{\includegraphics[scale=0.4,align=c]{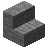}}
\newcommand{\mcstoneslab}{\includegraphics[scale=0.4,align=c]{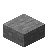}}
\newcommand{\mctorch}{\includegraphics[scale=0.4,align=c]{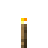}}
\newcommand{\mctrapdoor}{\includegraphics[scale=0.4,align=c]{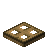}}
\newcommand{\mcwoodenpickaxe}{\includegraphics[scale=0.4,align=c]{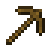}}
\newcommand{\mcwool}{\includegraphics[scale=0.4,align=c]{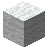}}

\newcommand{\mcheavypressureplate}{\includegraphics[scale=0.4,align=c]{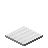}}
\newcommand{\mcironaxe}{\includegraphics[scale=0.4,align=c]{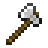}}
\newcommand{\mcironpickaxe}{\includegraphics[scale=0.4,align=c]{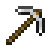}}
\newcommand{\mcironshovel}{\includegraphics[scale=0.4,align=c]{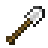}}
\newcommand{\mcironsword}{\includegraphics[scale=0.4,align=c]{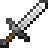}}
\newcommand{\mcirontrapdoor}{\includegraphics[scale=0.4,align=c]{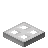}}
\newcommand{\mcstoneaxe}{\includegraphics[scale=0.4,align=c]{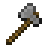}}
\newcommand{\mcstoneshovel}{\includegraphics[scale=0.4,align=c]{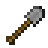}}
\newcommand{\mcstonesword}{\includegraphics[scale=0.4,align=c]{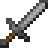}}
\newcommand{\mctripwirehook}{\includegraphics[scale=0.4,align=c]{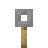}}
\newcommand{\mcwoodenaxe}{\includegraphics[scale=0.4,align=c]{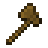}}
\newcommand{\mcwoodenshovel}{\includegraphics[scale=0.4,align=c]{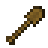}}
\newcommand{\mcwoodensword}{\includegraphics[scale=0.4,align=c]{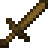}}
\newcommand{\mcironore}{\includegraphics[scale=0.4,align=c]{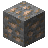}}
\newcommand{\mcdiamond}{\includegraphics[scale=0.4,align=c]{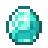}}
\newcommand{\mcdirt}{\includegraphics[scale=0.4,align=c]{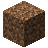}}

\title{Skill Reinforcement Learning and Planning for Open-World Long-Horizon Tasks}

\author{%
  Haoqi Yuan$^1$, Chi Zhang$^2$, Hongcheng Wang$^{1,4}$, Feiyang Xie$^3$, \\
  \textbf{Penglin Cai$^3$, Hao Dong$^{1}$, Zongqing Lu$^{1,4}$}\thanks{Correspondence to Zongqing Lu <zongqing.lu@pku.edu.cn>, Haoqi Yuan <yhq@pku.edu.cn>} \\  \\
   $^1$School of Computer Science, Peking University \\
   $^2$School of EECS, Peking University \\
   $^3$Yuanpei College, Peking University \\
   $^4$Beijing Academy of Artificial Intelligence
}

\begin{document}

\maketitle

\begin{abstract}
We study building multi-task agents in open-world environments. Without human demonstrations, learning to accomplish long-horizon tasks in a large open-world environment with reinforcement learning (RL) is extremely inefficient. To tackle this challenge, we convert the multi-task learning problem into learning basic skills and planning over the skills. Using the popular open-world game Minecraft as the testbed, we propose three types of fine-grained basic skills, and use RL with intrinsic rewards to acquire skills. A novel Finding-skill that performs exploration to find diverse items provides better initialization for other skills, improving the sample efficiency for skill learning. In skill planning, we leverage the prior knowledge in Large Language Models to find the relationships between skills and build a skill graph. When the agent is solving a task, our skill search algorithm walks on the skill graph and generates the proper skill plans for the agent. In experiments, our method accomplishes 40 diverse Minecraft tasks, where many tasks require sequentially executing for more than 10 skills. Our method outperforms baselines by a large margin and is the most sample-efficient demonstration-free RL method to solve Minecraft Tech Tree tasks. The project's website and code can be found at \href{https://sites.google.com/view/plan4mc}{https://sites.google.com/view/plan4mc}.
\end{abstract}

\section{Introduction}
Learning diverse tasks in open-ended worlds is a significant milestone toward building generally capable agents. 
Recent studies in multi-task reinforcement learning (RL) have achieved great success in many narrow domains like games~\citep{muzero} and robotics~\citep{meta-world}. 
However, transferring prior methods to open-world domains~\citep{open-ended-learning,minedojo} remains unexplored. Minecraft, a popular open-world game with an infinitely large world size and a huge variety of tasks, has been regarded as a challenging benchmark~\citep{minerl,minedojo}. 

Previous works usually build policies in Minecraft upon imitation learning, which requires expert demonstrations~\citep{minerl, goal-aware-learning, deps} or large-scale video datasets~\citep{vpt}.
Without demonstrations, RL in Minecraft is extremely sample-inefficient. A state-of-the-art model-based method~\citep{dreamer-v3} takes over 10M environmental steps to harvest cobblestones \mccobblestone, even if the block breaking speed of the game simulator is set to very fast additionally. 
This difficulty comes from at least two aspects. First, the world size is too large and the requisite resources are distributed far away from the agent. 
With partially observed visual input, the agent cannot identify its state or do effective exploration easily. Second, a task in Minecraft usually has a long horizon, with many sub-goals. For example, mining a cobblestone involves more than 10 sub-goals (from harvesting logs \mclog \ to crafting wooden pickaxes \mcwoodenpickaxe) and requires thousands of environmental steps. 

To mitigate the issue of learning long-horizon tasks, we propose to solve diverse tasks in a hierarchical fashion. In Minecraft, we define a set of basic skills. Then, solving a task can be decomposed into planning for a proper sequence of basic skills and executing the skills interactively. We train RL agents to acquire skills and build a high-level planner upon the skills.

\begin{figure}[!t]
  \centering
  \includegraphics[scale=0.7, trim={0cm, 12cm, 14.5cm, 0cm}, clip]{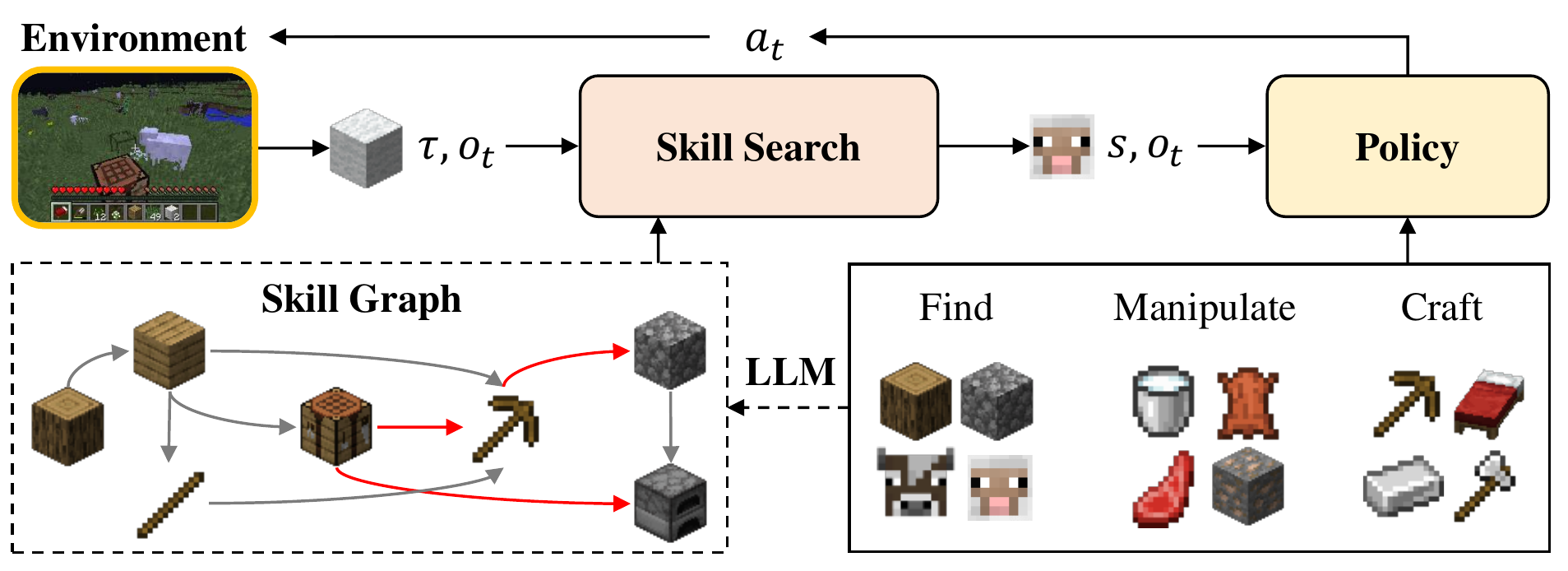}
  %\fbox{\rule[-.5cm]{0cm}{4cm} \rule[-.5cm]{4cm}{0cm}}
  \caption{Overview of \textbf{Plan4MC}. We categorize the basic skills in Minecraft into three types: Finding-skills, Manipulation-skills, and Crafting-skills. We train policies to acquire skills with reinforcement learning. With the help of LLM, we extract relationships between skills and construct a skill graph in advance, as shown in the dashed box. During online planning, the skill search algorithm walks on the pre-generated graph, decomposes the task into an executable skill sequence, and interactively selects policies to solve complex tasks.}
  \label{fig:pipeline}
\end{figure}

We find that training skills with RL remains challenging due to the difficulty in finding the required resources in the vast world. As an example, if we use RL to train the skill of harvesting logs, the agent can always receive 0 reward through random exploration since it cannot find a tree nearby. On the contrary, if a tree is always initialized close to the agent, the skill can be learned efficiently (Table~\ref{tab:rl-vs-imitation}). Thus, we propose to learn a Finding-skill that performs exploration to find items in the world and provides better initialization for all other skills, improving the sample efficiency of learning skills with RL. The Finding-skill is implemented with a hierarchical policy, maximizing the area traversed by the agent.

We split the skills in the recent work~\citep{deps} into more fine-grained basic skills and classify them into three types: Finding-skills, Manipulation-skills, and Crafting skills. 
%Unlike the previous definition of skills~\citep{deps}, we define more fine-grained basic skills and classify them into three types. 
Each basic skill solves an atomic task that may not be further divided. Such tasks have a shorter horizon and require exploration in smaller regions of the world. 
Thus, using RL to learn these basic skills is more feasible. 
To improve the sample efficiency of RL, we introduce intrinsic rewards to train policies for different types of skills.

For high-level skill planning, recent works~\citep{saycan, deps, voyager} demonstrate promising results via interacting with Large Language Models (LLMs). Though LLMs generalize to open-ended environments well and produce reasonable skill sequences, fixing their uncontrollable mistakes requires careful prompt engineering~\citep{inner-monologue, deps}. To make more flawless skill plans, we propose a complementary skill search approach. In the preprocessing stage, we use an LLM to generate the relationships between skills and construct a skill dependency graph. Then, given any task and the agent's condition (e.g., available resources/tools), we propose a search algorithm to interactively plan for the skill sequence. Figure \ref{fig:pipeline} illustrates our proposed framework, \textbf{Plan4MC}.

In experiments, we build 40 diverse tasks in the MineDojo~\citep{minedojo} simulator. These tasks involve executing diverse skills, including collecting basic materials \mclog \mccobblestone, crafting useful items \mccraftingtable \mcfurnace \mcwoodenpickaxe, and interacting with mobs \mccow \mcsheep. Each task requires planning and execution for 2\textasciitilde30 basic skills and takes thousands of environmental steps. Results show that Plan4MC accomplishes all the tasks and outperforms the baselines significantly. Also, Plan4MC can craft iron pickaxes \mcironpickaxe \ in the Minecraft Tech Tree and is much more sample-efficient than existing demonstration-free RL methods.

To summarize, our main contributions are:
\begin{itemize}
    \item To enable RL methods to efficiently solve diverse open-world tasks, we propose to learn fine-grained basic skills including a Finding-skill and train RL policies with intrinsic rewards. Thus, solving long-horizon tasks is transformed into planning over basic skills.
    \item Unlike previous LLM-based planning methods, we propose the skill graph and the skill search algorithm for interactive planning. The LLM only assists in the generation of the skill graph before task execution, avoiding uncontrollable failures caused by the LLM.  %The LLM generates skill relationships in advance, which can be easily checked and corrected manually, to avoid uncontrollable mistakes of the LLM.
    \item Our hierarchical agent achieves promising performance in diverse and long-horizon Minecraft tasks, demonstrating the great potential of using RL to build multi-task agents in open-ended worlds.
\end{itemize}

\section{Preliminaries}

\subsection{Problem Formulation}
In Minecraft, a task $\tau=(g,I)$ is defined with the combination of a goal $g$ and the agent's initial condition $I$, where $g$ represents the target entity to acquire in the task and $I$ represents the initial tools and conditions provided for the agent. For example, a task can be `harvest cooked\_beef \mccookedbeef \ with sword \mcdiamondsword \ in plains'. We model the task as a partially observable Markov decision process (POMDP)~\citep{pomdp}. $I$ determines the environment's initial state distribution. At each timestep $t$, the agent obtains the partial observation $o_t$, takes an action $a_t$ following its policy $\pi(a_t|o_{0:t}, \tau)$, and receives a sparse reward $r_t$ indicating task completion. The agent aims to maximize its expected return $R=\mathbb{E}_\pi \sum_t \gamma^tr_t$.

To solve complex tasks, humans acquire and reuse skills in the world, rather than learn each task independently from scratch. Similarly, to solve the aforementioned task, the agent can sequentially use the skills: harvest log \mclog, ..., craft furnace \mcfurnace, harvest beef \mcbeef, place furnace \mcfurnace, and craft cooked\_beef \mccookedbeef. 
Each skill solves a simple sub-task in a shorter time horizon, with the necessary tools and conditions provided. For example, the skill `craft cooked\_beef \mccookedbeef' solves the task `harvest cooked\_beef \mccookedbeef \ with beef \mcbeef, log \mclog, and placed furnace \mcfurnace'. 
Once the agent acquires an abundant set of skills $S$, it can solve any complex task by decomposing it into a sequence of sub-tasks and executing the skills in order. Meanwhile, by reusing a skill to solve different tasks, the agent is much better in memory and learning efficiency.

To this end, we convert the goal of solving diverse and long-horizon tasks in Minecraft into building a hierarchical agent. At the low level, we train policies $\pi_s$ to learn all the skills $s\in S$, where $\pi_s$ takes as input the RGB image and some auxiliary information (compass, location, biome, etc.), then outputs an action. At the high level, we study planning methods to convert a task $\tau$ into a skill sequence $(s_{\tau,1}, s_{\tau,2}, \cdots)$.

\subsection{Skills in Minecraft}

Recent works mainly rely on imitation learning to learn Minecraft skills efficiently. In MineRL competition~\citep{minerl2021}, a human gameplay dataset is accessible along with the Minecraft environment. All of the top methods in competition use imitation learning to some degree, to learn useful behaviors in limited interactions. In VPT~\citep{vpt}, a large policy model is pre-trained on a massive labeled dataset using behavior cloning. By fine-tuning on smaller datasets, policies are acquired for diverse skills. 

However, without demonstration datasets, learning Minecraft skills with reinforcement learning (RL) is difficult. MineAgent~\citep{minedojo} shows that PPO~\citep{ppo} can only learn a small set of skills. PPO with sparse reward fails in `milk a cow' and `shear a sheep', though the distance between target mobs and the agent is set within 10 blocks. We argue that with the high dimensional state and action space, open-ended large world, and partial observation, exploration in Minecraft tasks is extremely difficult.

We conduct a study for RL to learn skills with different difficulties in Table \ref{tab:rl-vs-imitation}. We observe that RL has comparable performance to imitation learning only when the task-relevant entities are initialized very close to the agent. Otherwise, RL performance decreases significantly. This motivates us to further divide skills into fine-grained skills. We propose a \textbf{Finding-skill} to provide a good initialization for other skills. For example, the skill of `milk a cow' is decomposed into `find a cow' and `harvest milk\_bucket'. After finding a cow nearby, `harvest milk\_bucket' can be accomplished by RL with acceptable sample efficiency. Thus, learning such fine-grained skills is easier for RL, and they together can still accomplish the original task.

\begin{table}[!t]
  \caption{Minecraft skill performance of imitation learning (behavior cloning with MineCLIP backbone, reported in~\citep{goal-aware-learning}) versus reinforcement learning. \textit{Better init.} means target entities are closer to the agent at initialization. The RL method for each task is trained with proper intrinsic rewards. All RL results are averaged on the last 100 training epochs and 3 training seeds.}
  \label{tab:rl-vs-imitation}
  \centering
  \begin{tabular}{lccccc}
    \toprule
    Skill     &    \mcmilkbucket   & \mcwool  &  \mccow &  \mcsheep & \mclog \\
    \midrule
    Behavior Cloning & --  & -- & 0.25 & 0.27 & 0.16 \\
    RL  & 0.40$\pm$0.20 & 0.26$\pm$0.22 & 0.04$\pm$0.02 & 0.04$\pm$0.01 & 0.00$\pm$0.00 \\
    RL (\textit{better init.}) & 0.99$\pm$0.01 & 0.81$\pm$0.02 & 0.16$\pm$0.06 & 0.14$\pm$0.07 & 0.44$\pm$0.10 \\
    \bottomrule
  \end{tabular}
\end{table}

%\subsection{Skill Planning for Hard Tasks}
%Previous works use LLM prompting xxxx, based on the strong assumption that LLM contains expert knowledge about the specific domain. LLM fails xxx. XXX propose method on grounding LLM, interactive planning to increase success rate. These methods are complicated and require human engineering.  xxx

\section{Learning Basic Skills with Reinforcement Learning}
Based on the discussion above, we propose three types of fine-grained basic skills, which can compose all Minecraft tasks.
\begin{itemize}
    \item Finding-skills: starts from any location, the agent explores to find a target and approaches the target. The target can be any block or entity that exists in the world.
    \item Manipulation-skills: given proper tools and the target in sight, the agent interacts with the target to obtain materials. These skills include diverse behaviors, like mining ores, killing mobs, and placing blocks.
    \item Crafting-skills: with requisite materials in the inventory and crafting table or furnace placed nearby, the agent crafts advanced materials or tools.
\end{itemize}

%Except for crafting-skills that can be executed with only a single action, we study learning basic skills with RL.

\subsection{Learning to Find with a Hierarchical Policy}
Finding items is a long-horizon difficult task for RL. To find an unseen tree on the plains, the agent should take thousands of steps to explore the world map as much as possible. A random policy fails to do such exploration, as shown in Appendix \ref{appendix:find-skill}.
Also, it is too costly to train different policies for various target items. To simplify this problem, considering to explore on the world's surface only, we propose to train a target-free hierarchical policy to solve all the Finding-skills.

\begin{figure}[!t]
  \centering
  \includegraphics[scale=0.7, trim={0cm, 14.5cm, 14.5cm, 0cm}, clip]{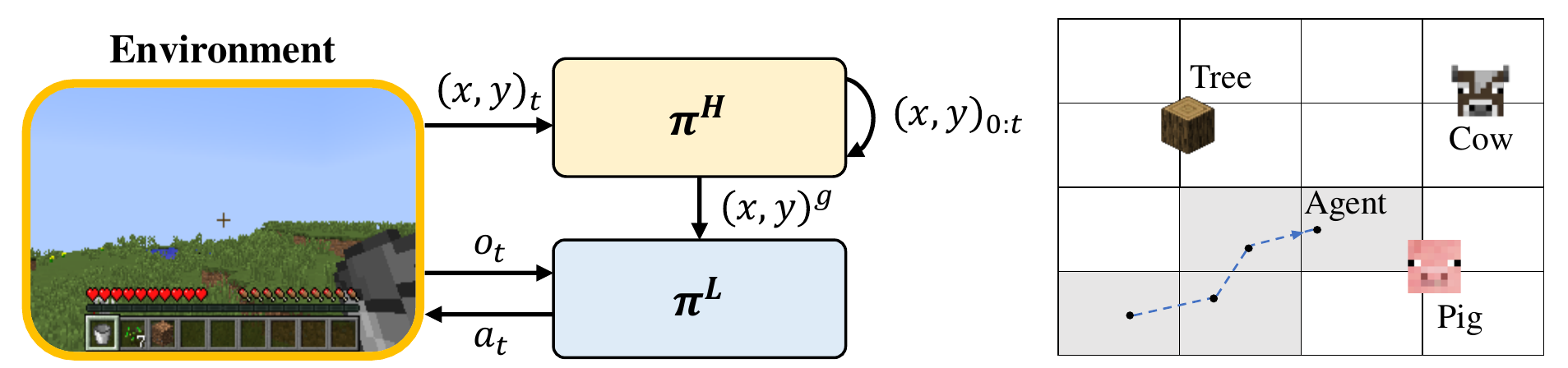}
  %\fbox{\rule[-.5cm]{0cm}{4cm} \rule[-.5cm]{4cm}{0cm}}
  \caption{The proposed hierarchical policy for Finding-skills. The high-level recurrent policy $\pi^H$ observes historical positions $(x,y)_{0:t}$ from the environment and generates a goal position $(x,y)^g$. The low-level policy $\pi^L$ is a goal-based policy to reach the goal position. The right figure shows a top view of the agent's exploration trajectory, where the walking paths of the low-level policy are shown in blue dotted lines, and the goal is changed by the high-level policy at each black spot. The high-level policy is optimized to maximize the state count in the grid world, which is shown in the grey background.}
  \label{fig:findskill}
\end{figure}

Figure \ref{fig:findskill} demonstrates the hierarchical policy for Finding-skills.
The high-level policy $\pi^H\left((x,y)^g | (x,y)_{0:t}\right)$ observes historical locations $(x,y)_{0:t}$ of the agent, and outputs a goal location $(x,y)^g$. It drives the low-level policy $\pi^L\left(a_t|o_t,(x,y)^g\right)$ to reach the goal location. We assume that target items are uniformly distributed on the world's surface. To maximize the chance to find diverse targets, the objective for the high-level policy is to maximize its reached area. We divide the world's surface into discrete grids, where each grid represents a $10\times 10$ area. We use state count in the grids as the reward for the high-level policy. The low-level policy obtains the environmental observation $o_t$ and the goal location $(x,y)^g$ proposed by the high-level policy, and outputs an action $a_t$. We reward the low-level policy with the distance change to the goal location. 

To train the hierarchical policy with acceptable sample complexity, we pre-train the low-level policy with randomly generated goal locations using DQN~\citep{dqn}, then train the high-level policy using PPO~\citep{ppo} with the fixed low-level policy. During test, to find a specific item, the agent first explores the world with the hierarchical policy until a target item is detected in its lidar observations. Then, the agent executes the low-level policy conditioned on the detected target's location, to reach the target item. Though we use additional lidar information here, we believe that without this information, we can also implement the success detector for Finding-skills with computer vision models~\citep{vlm-detectors}.

\subsection{Manipulation and Crafting}
\label{subsec:manipulation-skills}
By executing the pre-trained Finding-skills, we can instantiate the manipulation tasks with requisite target items nearby, making the manipulation tasks much easier.
%Manipulation tasks are instantiated with requisite tools in the inventory and target items nearby. 
To train the Manipulation-skills in Minecraft, we can either make a training environment with the target item initialized nearby or run the Finding-skills to reach a target item.
For example, to train the skill `harvest milk\_bucket \mcmilkbucket', we can either spawn a cow \mccow \ close to the agent using the Minecraft built-in commands, or execute the Finding-skills until a cow is reached. %For `harvest log \mclog', we spawn the agent in the forest to make sure nearby trees exist. Thus, Manipulation-skills can be accomplished in fewer environmental steps.
The latter is similar in the idea to Go-Explore~\citep{go-explore}, and is more suitable for other environments that do not have commands to initialize the target items nearby.

We adopt MineCLIP~\citep{minedojo} to guide the agent with intrinsic rewards. The pre-trained MineCLIP model computes the CLIP reward based on the similarity between environmental observations (frames) and the language descriptions of the skill. We train the agent using PPO with self-imitation learning, to maximize a weighted sum of intrinsic rewards and extrinsic success (sparse) reward. 
Details for training basic skills can be found in Appendix \ref{appendix:skill-details}. 

For the Crafting-skills, they can be executed with only a single action in MineDojo~\citep{minedojo}.

% \begin{figure}
%   \centering
%   \fbox{\rule[-.5cm]{0cm}{4cm} \rule[-.5cm]{4cm}{0cm}}
%   \caption{Sample figure caption.}
% \end{figure}

% \begin{table}
%   \caption{Sample table title}
%   \label{sample-table}
%   \centering
%   \begin{tabular}{lll}
%     \toprule
%     \multicolumn{2}{c}{Part}                   \\
%     \cmidrule(r){1-2}
%     Name     & Description     & Size ($\mu$m) \\
%     \midrule
%     Dendrite & Input terminal  & $\sim$100     \\
%     Axon     & Output terminal & $\sim$10      \\
%     Soma     & Cell body       & up to $10^6$  \\
%     \bottomrule
%   \end{tabular}
% \end{table}

\section{Solving Minecraft Tasks via Skill Planning}
In this section, we present our skill planning method for solving diverse hard tasks. A skill graph is generated in advance with a Large Language Model (LLM), enabling searching for correct skill sequences on the fly.

\definecolor{backcolour}{rgb}{0.95,0.95,0.95}
\lstset{
    backgroundcolor=\color{backcolour},   
    basicstyle=\ttfamily\footnotesize,
    %breakatwhitespace=false,         
    breaklines=true,                 
    captionpos=t,                   
    %keepspaces=true,
    %showspaces=false,                
    %showstringspaces=false,
    %showtabs=false,                  
    %tabsize=2,
    escapeinside={``}
}

\subsection{Constructing Skill Graph with Large Language Models}
A correct plan  $(s_{\tau,1}, s_{\tau,2}, \cdots)$ for a task $\tau=(g,I)$ should satisfy two conditions. (1) For each $i$, $s_{\tau,i}$ is executable after $(s_{\tau,1}, \cdots, s_{\tau,i-1})$ are accomplished sequentially with initial condition $I$. (2) The target item $g$ is obtained after all the skills are accomplished sequentially, given initial condition $I$. To enable searching for such plans, we should be able to verify whether a plan is correct. Thus, we should know what condition is required and what is obtained for each skill. We define such information of skills in a structured format. As an example, information for skill `crafting stone\_pickaxe \mcstonepickaxe' is:
\begin{lstlisting}
stone_pickaxe {consume: {cobblestone: 3, stick: 2}, 
require: {crafting_table_nearby: 1}, obtain: {stone_pickaxe: 1}}
\end{lstlisting}
Each item in this format is also a skill. Regarding them as graph nodes, this format shows a graph structure between skill `stone\_pickaxe' and skills `cobblestone', `stick', `crafting\_table\_nearby'. The directed edge from `cobblestone` to `stone\_pickaxe' is represented as (3, 1, consume), showing the quantity relationship between parent and child, and that the parent item will be consumed during skill execution. In fact, in this format, all the basic skills in Minecraft construct a large directed acyclic graph with hundreds of nodes. The dashed box in Figure \ref{fig:pipeline} shows a small part of this graph, where grey arrows denote `consume' and red arrows denote `require'.

To construct the skill graph, we generate structured information for all the skills by interacting with ChatGPT (GPT-3.5)~\citep{gpt}, a high-performance LLM. Since LLMs are trained on large-scale internet datasets, they obtain rich knowledge in the popular game Minecraft. In prompt, we give a few demonstrations and explanations about the format, then ask ChatGPT to generate other skills information. %We find that ChatGPT makes few mistakes, and these mistakes can be easily detected and fixed by human prior. 
Dialog with ChatGPT can be found in Appendix \ref{appendix:llmprompt}. 

\subsection{Skill Search Algorithm}

Our skill planning method is a depth-first search (DFS) algorithm on the skill graph. Given a task $\tau=(g,I)$, we start from the node $g$ and do DFS toward its parents, opposite to the edge directions. In this process, we maintain all the possessing items starting from $I$. Once conditions for the skill are satisfied or the skill node has no parent, we append this skill into the planned skill list and modify the maintained items according to the skill information. The resulting skill list is ensured to be executable and target-reaching. %Algorithm \ref{alg:dfs} presents the pseudocode for the skill search algorithm.

To solve a long-horizon task, since the learned low-level skills are possible to fail, we alternate skill planning and skill execution until the episode terminates. After each skill execution, we update the agent's condition $I'$ based on its inventory and the last executed skill, and search for the next skill with $\tau'=(g,I')$. %We summarize this testing process in Algorithm \ref{alg:solve-task}.

We present the pseudocode for the skill search algorithm and the testing process in Appendix \ref{appendix:algorithms}.

\section{Experiments}
In this section, we evaluate and analyze our method with baselines and ablations in challenging Minecraft tasks. Section \ref{subsec:pretrain} introduces the implementation of basic skills. In Section \ref{subsec:tasks}, we introduce the setup for our evaluation task suite. In Section \ref{subsec:skill-learning} and \ref{subsec:skill-planning}, we present the experimental results and analyze skill learning and planning respectively.

\subsection{Training Basic Skills}
\label{subsec:pretrain}
To pre-train basic skills with RL, we use the environments of programmatic tasks in MineDojo~\citep{minedojo}. 
To train Manipulation-skills, for simplicity, we specify the environment that initializes target mobs or resources close to the agent. For the Go-Explore-like training method without specified environments discussed in Section \ref{subsec:manipulation-skills}, we present the results in Appendix \ref{appendix:go-explore}, which does not underperform the former.

For Manipulation-skills and the low-level policy of Finding-skills, we adopt the policy architecture of MineAgent~\citep{minedojo}, which uses a fixed pre-trained MineCLIP image encoder and processes features using MLPs. To explore in a compact action space, we compress the original large action space into $12\times 3$ discrete actions. For the high-level policy of Finding-skills, which observes the agent's past locations, we use an LSTM policy and train it with truncated BPTT~\citep{truncated-bptt}. We pick the model with the highest success rate on the smoothed training curve for each skill, and fix these policies in all tasks. Implementation details can be found in Appendix \ref{appendix:skill-details}. 

Note that Plan4MC totally takes 7M environmental steps in training, and can unlock the iron pickaxe \mcironpickaxe \ in the Minecraft Tech Tree in test. The sample efficiency greatly outperforms all other existing demonstration-free RL methods~\citep{dreamer-v3, vpt}.

\subsection{Task Setup}
\label{subsec:tasks}

Based on MineDojo~\citep{minedojo} programmatic tasks, we set up an evaluation benchmark consisting of four groups of diverse tasks: cutting trees \mclog \ to craft primary items, mining cobblestones \mccobblestone \ to craft intermediate items, mining iron ores \mcironore \ to craft advanced items, and interacting with mobs \mccow \ to harvest food and materials. Each task set has 10 tasks, adding up to a total of 40 tasks. With our settings of basic skills, these tasks require 25 planning steps on average and maximally 121 planning steps. We estimate the number of the required steps for each task with the sum of the steps of the initially planned skills and double this number to be the maximum episode length for the task, allowing skill executions to fail. The easiest tasks have 3000 maximum steps, while the hardest tasks have 12000. More details about task setup are listed in Appendix \ref{appendix:task-setup}. To evaluate the success rate on each task, we average the results over 30 test episodes.

\subsection{Skill Learning}
\label{subsec:skill-learning}

We first analyze learning basic skills. While we propose three types of fine-grained basic skills, others directly learn more complicated and long-horizon skills. We introduce two baselines to study learning skills with RL.

\textbf{MineAgent~\citep{minedojo}.} Without decomposing tasks into basic skills, MineAgent solves tasks using PPO and self-imitation learning with the CLIP reward. For fairness, we train MineAgent in the test environment for each task. The training takes 7M environmental steps, which is equal to the sum of environmental steps we take for training all the basic skills. We average the success rate of trajectories in the last 100 training epochs (around 1M environment steps) to be its test success rate. Since MineAgent has no actions for crafting items, we hardcode the crafting actions into the training code. During trajectory collection, at each time step where the skill search algorithm returns a Crafting-skill, the corresponding crafting action will be executed. Note that, if we expand the action space for MineAgent rather than automatically execute crafting actions, the exploration will be much harder.

\textbf{Plan4MC w/o Find-skill.} None of the previous work decomposes a skill into executing Finding-skills and Manipulation-skills. Instead, finding items and manipulations are done with a single skill. Plan4MC w/o Find-skill implements such a method. It skips all the Finding-skills in the skill plans during test. Manipulation-skills take over the whole process of finding items and manipulating them. 

\begin{table}[!t]
  \caption{Average success rates on four task sets of our method, all the baselines and ablation methods. Success rates on all the single tasks are listed in Appendix \ref{appendix:results-all}.}
  \label{tab:results}
  \centering
  \begin{tabular}{lcccc}
    \toprule
    Task Set  & Cut-Trees & Mine-Stones &  Mine-Ores & Interact-Mobs \\
    \midrule
    MineAgent  & 0.003  & 0.026 & 0.000 & 0.171 \\
    {Plan4MC w/o Find-skill}  & 0.187  & 0.097 & 0.243 & 0.170 \\
    \midrule
    {Interactive LLM}  &  0.260 & 0.067 & 0.030 & 0.247  \\
    {Plan4MC Zero-shot}  & 0.183  &  0.000 & 0.000 & 0.133 \\
    {Plan4MC 1/2-steps}  & 0.337  &  0.163 & 0.143 & 0.277 \\
    \midrule
    \textbf{Plan4MC}  & \textbf{0.417} &  \textbf{0.293} & \textbf{0.267} & \textbf{0.320} \\
    \bottomrule
  \end{tabular}
\end{table}

Table \ref{tab:results} shows the test results for these methods. Plan4MC outperforms two baselines on the four task sets. MineAgent fails on the task sets of Cut-Trees, Mine-Stones and Mine-Ores, since taking many attacking actions continually to mine blocks in Minecraft is an exploration difficulty for RL on long-horizon tasks. On the contrary, MineAgent achieves performance comparable to Plan4MC's on some easier tasks \mcmilkbucket\mcwool\mcbeef\mcmutton \ in Interact-Mobs, which requires fewer environmental steps and planning steps. Plan4MC w/o Find-skill consistently underperforms Plan4MC on all the tasks, showing that introducing Finding-skills is beneficial for solving hard tasks with basic skills trained by RL. Because there is no Finding-skill in harvesting iron ores, their performance gap on Mine-Ores tasks is small.

\begin{figure}[!t]
  \centering
  \includegraphics[scale=0.335, trim={4.5cm, 0cm, 4.5cm, 0.5cm}, clip]{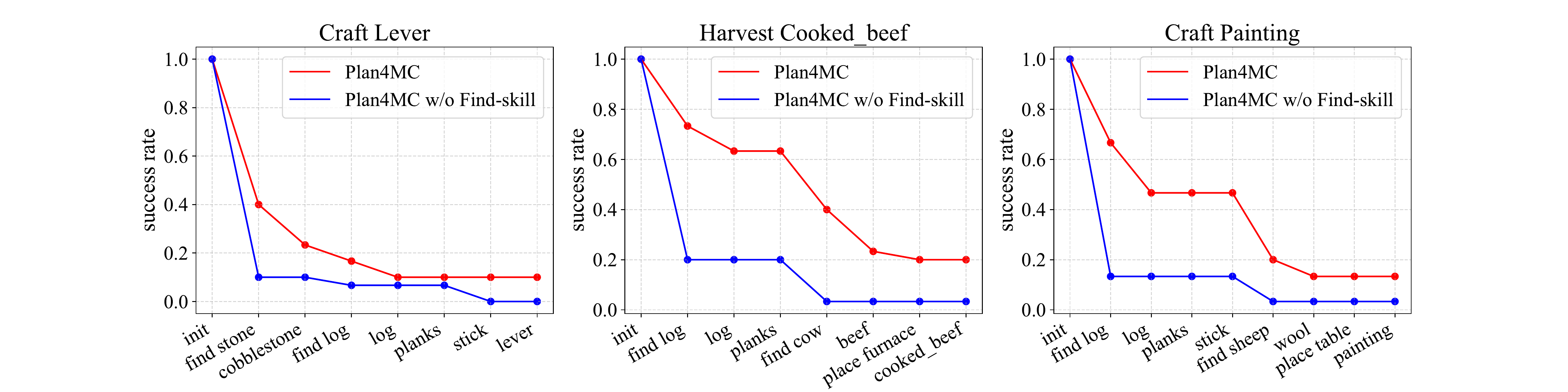}
  %\fbox{\rule[-.5cm]{0cm}{4cm} \rule[-.5cm]{4cm}{0cm}}
  \caption{Success rates of Plan4MC with/without Finding-skills at each skill planning step, on three long-horizon tasks. We arrange the initially planned skill sequence on the horizontal axis and remove the repeated skills. The success rate of each skill represents the probability of successfully executing this skill at least once in a test episode. Specifically, the success rate is always 1 at task initialization, and the success rate of the last skill is equal to the task's success rate.}
  \label{fig:ablation-find}
\end{figure}

To further study Finding-skills, we present the success rate at each planning step in Figure \ref{fig:ablation-find} for three tasks. The curves of Plan4MC and Plan4MC w/o Find-skill have large drops at Finding-skills. Especially, the success rates at finding cobblestones and logs decrease the most, because these items are harder to find in the environment compared to mobs. In these tasks, we compute the average success rate of Manipulation-Skills, conditioned on the skill before the last Finding-skills being accomplished. While Plan4MC has a conditional success rate of 0.40, Plan4MC w/o Find-skill decreases to 0.25, showing that solving sub-tasks with additional Finding-skills is more effective.

As shown in Table \ref{tab:skill-success-rate}, most Manipulation-skills have slightly lower success rates in test than in training, due to the domain gap between test and training environments. However, this decrease does not occur in skills \mcmilkbucket\mcwool \ that are trained with a large initial distance of mobs/items, as pre-executed Finding-skills provide better initialization for Manipulation-skills during the test and thus the success rate may increase. In contrast, the success rates in the test without Finding-skills are significantly lower.

\begin{table}[!t]
  \caption{Success rates of Manipulation-skills in training and test. \textit{Training init. distance} is the maximum distance for mobs/items initialization in training skills. Note that in test, executing Finding-skills will reach the target items within a distance of 3. \textit{Training success rate} is averaged over 100 training epochs around the selected model's epoch. \textit{Test success rate} is computed from the test rollouts of all the tasks, while \textit{w/o Find} refers to Plan4MC w/o Find-skill.}
  \label{tab:skill-success-rate}
  \centering
  \begin{tabular}{lccccccc}
    \toprule
    Manipulation-skills & Place & \mcmilkbucket & \mcwool & \mcbeef & \mcmutton & \mclog & \mccobblestone \\ \specialrule{0em}{0pt}{3pt}
    Training init. distance &  \--\-- &  10  &  10 &  2  &  2  &  \--\-- &  \--\-- \\
        \midrule
    Training success rate &  0.98 & 0.50  & 0.27 & 0.21  & 0.30 &  0.56 & 0.47  \\
    Test success rate & 0.77  & 0.71 &  0.26 &  0.27 & 0.16  &  0.33 & 0.26  \\
    Test success rate (w/o Find) & 0.79 & 0.07 & 0.03  &  0.03 &  0.02 & 0.05 & 0.06  \\
    \bottomrule
  \end{tabular}
\end{table}

% \begin{figure}[!t]
%   \centering
%   \includegraphics[scale=0.335, trim={4.5cm, 0cm, 4.5cm, 0.5cm}, clip]{fig/plansteps_plan2.pdf}
%   %\fbox{\rule[-.5cm]{0cm}{4cm} \rule[-.5cm]{4cm}{0cm}}
%   \caption{Success rates of Plan4MC compared with ablated skill planning methods at each planning step, on three long-horizon tasks. We arrange the initial planned skill sequence on the horizontal axis and remove the repeated skills. The success rate of each skill represents the probability of successfully executing this skill at least once in a test episode. Specifically, the success rate is always 1 at task initialization, and the success rate of the last skill is equal to the task's success rate.}
%   \label{fig:ablation-plan}
% \end{figure}

\subsection{Skill Planning}
\label{subsec:skill-planning}

For skill planning in open-ended worlds, recent works~\citep{translation-lm, inner-monologue, saycan, code-as-policy, deps} generate plans or sub-tasks with LLMs. We study these methods on our task sets and implement a best-performing baseline to compare with Plan4MC.

\textbf{Interactive LLM.} We implement an interactive planning baseline using LLMs. We take ChatGPT~\citep{gpt} as the planner, which proposes skill plans based on prompts including descriptions of tasks and observations. Similar to chain-of-thoughts prompting~\citep{chain-of-thoughts}, we provide few-shot demonstrations with explanations to the planner at the initial planning step. In addition, we add several rules for planning into the prompt to fix common errors that the model encountered during test. At each subsequent planning step, the planner will encounter one of the following cases: the proposed skill name is invalid, the skill is already done, skill execution succeeds, and skill execution fails. We carefully design language feedback for each case and ask the planner to re-plan based on inventory changes. For low-level skills, we use the same pre-trained skills as Plan4MC.

Also, we conduct ablations on our skill planning designs.

\textbf{Plan4MC Zero-shot.} This is a zero-shot variant of our interactive planning method, proposing a skill sequence at the beginning of each task only. The agent executes the planned skills sequentially until a skill fails or the environment terminates. This planner has no fault tolerance for skills execution.

\textbf{Plan4MC 1/2-steps.} In this ablation study, we half the test episode length and require the agent to solve tasks more efficiently.

Success rates for each method are listed in Table \ref{tab:results}. We find that Interactive LLM has comparable performance to Plan4MC on the task set of Interact-Mobs, where most tasks require less than 10 planning steps. In Mine-Stones and Mine-Ores tasks with long-horizon planning, the LLM planner is more likely to make mistakes, resulting in worse performance. The performance of Plan4MC Zero-shot is much worse than Plan4MC in all the tasks, since a success test episode requires accomplishing each skill in one trial. The decrease is related to the number of planning steps and skills success rates in Table \ref{tab:skill-success-rate}. Plan4MC 1/2-steps has the least performance decrease to Plan4MC, showing that Plan4MC can solve tasks in a more limited episode length.

%As shown in Figure \ref{fig:ablation-plan}, Plan4MC 1/2-steps has close performance to Plan4MC at each planning step, while the Plan4MC Zero-shot fails after executing skills with low success rates. Plan4MC 1/2-steps has the least performance decrease to Plan4MC, showing that Plan4MC can solve tasks in a very limited episode length.

% \subsection{Obtain Stone Pickaxe \mcstonepickaxe \ with Bare Hands}
% Obtaining stone pickaxe with bare hands is the \textit{most} challenging task in Minecraft tech tree with our basic skills, which involves 25 planning steps and 10 basic skills. Note that this task is not in the 24 benchmark tasks. Plan4MC achieves a success rate of 18\% in 100 trials, while all other methods fail with 0\% success rates. A test episode is shown in Figure \ref{fig:stonepickaxe}.

\section{Related Work}

\textbf{Minecraft.} In recent years, the open-ended world Minecraft has received wide attention in machine learning research. Malmo~\citep{malmo}, MineRL~\citep{minerl} and MineDojo~\citep{minedojo} build benchmark environments and datasets for Minecraft. Previous works in MineRL competition~\citep{minerl2019, minerl2020, minerl2021} study the ObtainDiamond task with hierarchical RL~\citep{minerl2019, hdqfd, seihai,juewumc} and imitation learning~\citep{imitation-minecraft, minerl2020}. Other works explore multi-task learning~\citep{h-drln, multitask-curriculum, goal-aware-learning, DECKARD}, unsupervised skill discovery~\citep{unsupervised-skill}, LLM-based planning~\citep{deps,voyager,gitm}, and pre-training from videos~\citep{vpt,steve1, minedojo, clip4mc}. Our work falls under reinforcement learning and planning in Minecraft.

\textbf{Learning Skills in Minecraft.} Acquiring skills is crucial for solving long-horizon tasks in Minecraft. Hierarchical approaches~\citep{seihai, juewumc} in MineRL competition learn low-level skills with imitation learning. VPT~\citep{vpt} labels internet-scale datasets and pre-trains a behavior-cloning agent to initialize for diverse tasks. Recent works~\citep{goal-aware-learning, deps, DECKARD} learn skills based on VPT. Without expert demonstrations, MineAgent~\citep{minedojo} and CLIP4MC~\citep{clip4mc} learn skills with RL and vision-language rewards. But they can only acquire a small set of skills. Unsupervised skill discovery~\citep{unsupervised-skill} learns skills that only produce different navigation behaviors. In our work, to enable RL to acquire diverse skills, we learn fine-grained basic skills with intrinsic rewards.

\textbf{Planning with Large Language Models.} With the rapid progress of LLMs~\citep{gpt, palm}, many works study LLMs as planners in open-ended worlds. To ground language models, SayCan~\citep{saycan} combines LLMs with skill affordances to produce feasible plans, Translation LMs~\citep{translation-lm} selects demonstrations to prompt LLMs, and LID~\citep{lid} finetunes language models with tokenized interaction data. Other works study interactive planning for error correction. Inner Monologue~\citep{inner-monologue} proposes environment feedback to the planner. DEPS~\citep{deps} introduces descriptor, explainer, and selector to generate plans by LLMs. In our work, we leverage the LLM to generate a skill graph and introduce a skill search algorithm to eliminate planning mistakes.

\section{Conclusion and Discussion}
In this paper, we propose a framework to solve diverse long-horizon open-world tasks with reinforcement learning and planning. 
To tackle the exploration and sample efficiency issues, we propose to learn fine-grained basic skills with RL and introduce a general Finding-skill to provide good environment initialization for skill learning.
In Minecraft, we design a graph-based planner, taking advantage of the prior knowledge in LLMs and the planning accuracy of the skill search algorithm.
%we propose Plan4MC to learn basic skills with reinforcement learning and plan for tasks with the skill search algorithm on the skill graph. 
Experiments on 40 challenging Minecraft tasks verify the advantages of Plan4MC over various baselines. 

Though we implement Plan4MC in Minecraft, our method is extendable to other similar open-world environments and draws insights on building multi-task learning systems. We leave the detailed discussion in Appendix \ref{appendix:discussion}.

A limitation of this work is that the Finding-skill is not aware of its goal during exploration, making the goal-reaching policy sub-optimal.  %cannot explore the underground world. 
Future work needs to improve its efficiency by training a goal-based policy. Moreover, if the LLM lacks domain knowledge, how to correct the LLM's outputs is a problem worth studying in the future. Providing documents and environmental feedback to the LLM is a promising direction.
%Moreover, this work pre-trains all the required skills for solving hard tasks. Acquiring skills through online interaction with an unspecified open-ended environment is also left as future work.

%%%%%%%%%%%%%%%%%%%%%%%%%%%%%%%%%%%%%%%%%%%%%%%%%%%%%%%%%%%%
\bibliographystyle{plain}
\bibliography{workshop}

%%%%%%%%%%%%%%%%%%%%%%%%%%%%%%%%%%%%%%%%%%%%%%%%%%%%%%%%%%%%

\newpage
\appendix
% & \mccraftingtable & \mcbowl & \mcchest & \mctrapdoor & \mcsign & \mcwoodenpickaxe & \mcmilkbucket & \mcwool & \mcbed & \mcbeef & \mcmutton

\section{The Necessity of Learning the Finding-skill}
\label{appendix:find-skill}

\begin{figure}[htbp]
  \centering
  \includegraphics[scale=0.12, trim={0cm, 0cm, 0cm, 0.5cm}, clip]{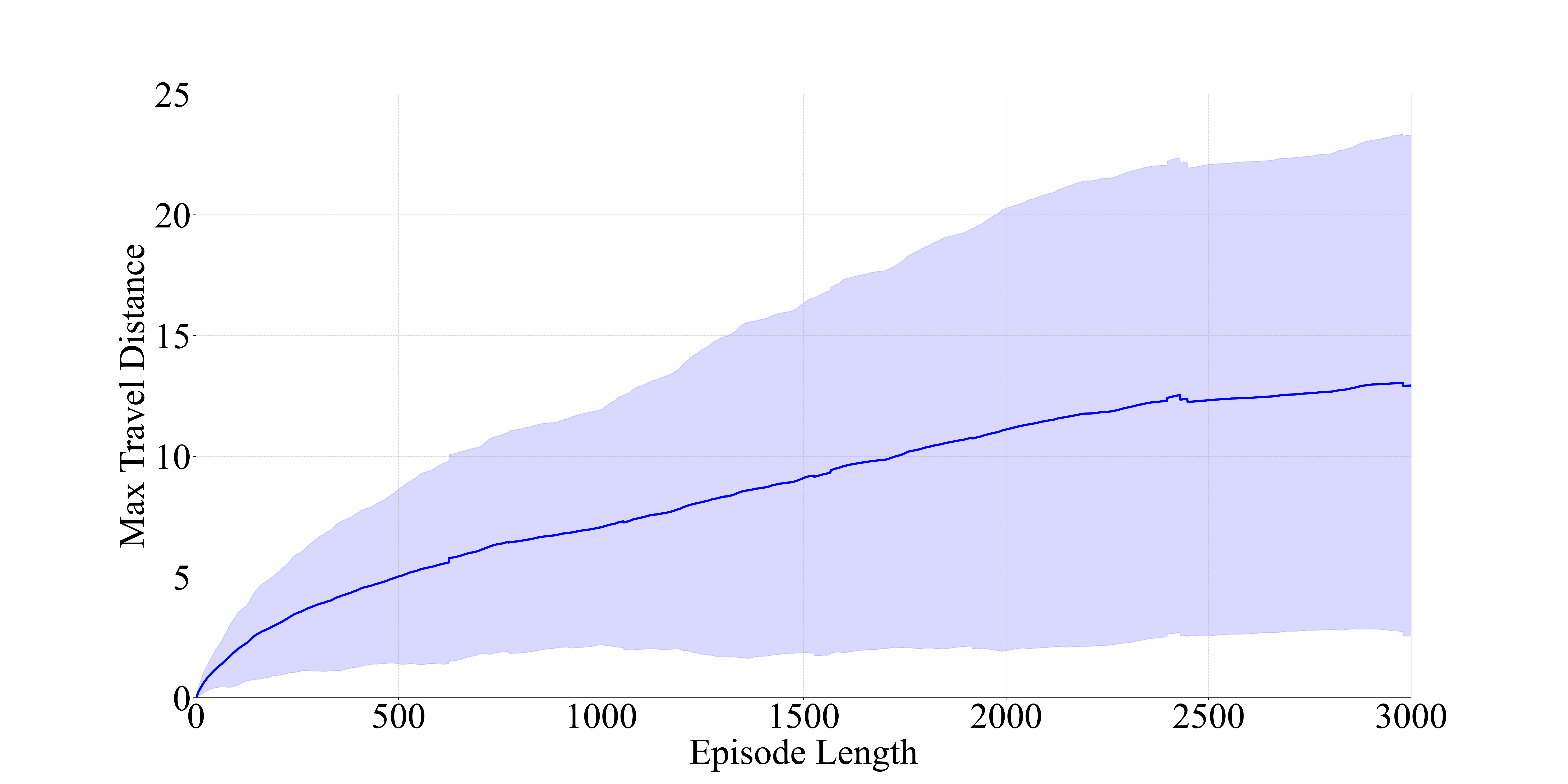}
  %\fbox{\rule[-.5cm]{0cm}{4cm} \rule[-.5cm]{4cm}{0cm}}
  \vspace{-2mm}
  \caption{Maximal travel distance to the spawning point a random policy could reach in Minecraft, under different episode lengths. We test for 100 episodes, with different randomly generated worlds and agent parameters. Note that all Manipulation-skills we trained have episode lengths less than 1000 to ensure sample efficiency.}
  \label{fig:random-travel}
\end{figure}

We demonstrate the exploration difficulty of learning skills in Minecraft. Figure \ref{fig:random-travel} shows that a random policy can only travel to a distance of 5 blocks on plains within 500 steps. Since trees are rare on the plains and usually have $>20$ distances to the player, an RL agent starting from a random policy can fail to collect logs on plains.

In Table \ref{tab:travel-dis}, we compare the travel distances of a random policy, a hand-coded walking policy, and our Finding-skill pre-trained with RL. We find that the Finding-skill has a stronger exploration ability than the other two policies.

\begin{table}[htbp]
  \caption{Maximal travel distance on plains of a random policy, a hand-coded policy which always takes forward+jump and randomly turns left or right, and our Finding-skill. }
  \label{tab:travel-dis}
  \centering
  \begin{tabular}{cccc}
    \toprule
    Episode length & 200 & 500 & 1000 \\
    \midrule
    Random Policy & $3.0 \pm 2.1$ & $5.0 \pm 3.6$ & $7.1 \pm 4.9$ \\
    Hand-coded Policy & $7.1 \pm 2.7$ & $11.7 \pm 4.4$ & $18.0 \pm 6.6$ \\
    Finding-skill & $12.6 \pm 5.6$ & $18.5 \pm 9.3$ & $25.7 \pm 12.1$ \\
    \bottomrule
  \end{tabular}
\end{table}

\newpage
\section{Pipeline Demonstration}

\begin{figure}[htbp]
  \centering
  \includegraphics[scale=0.43, trim={0cm, 0cm, 0cm, 0cm}, clip]{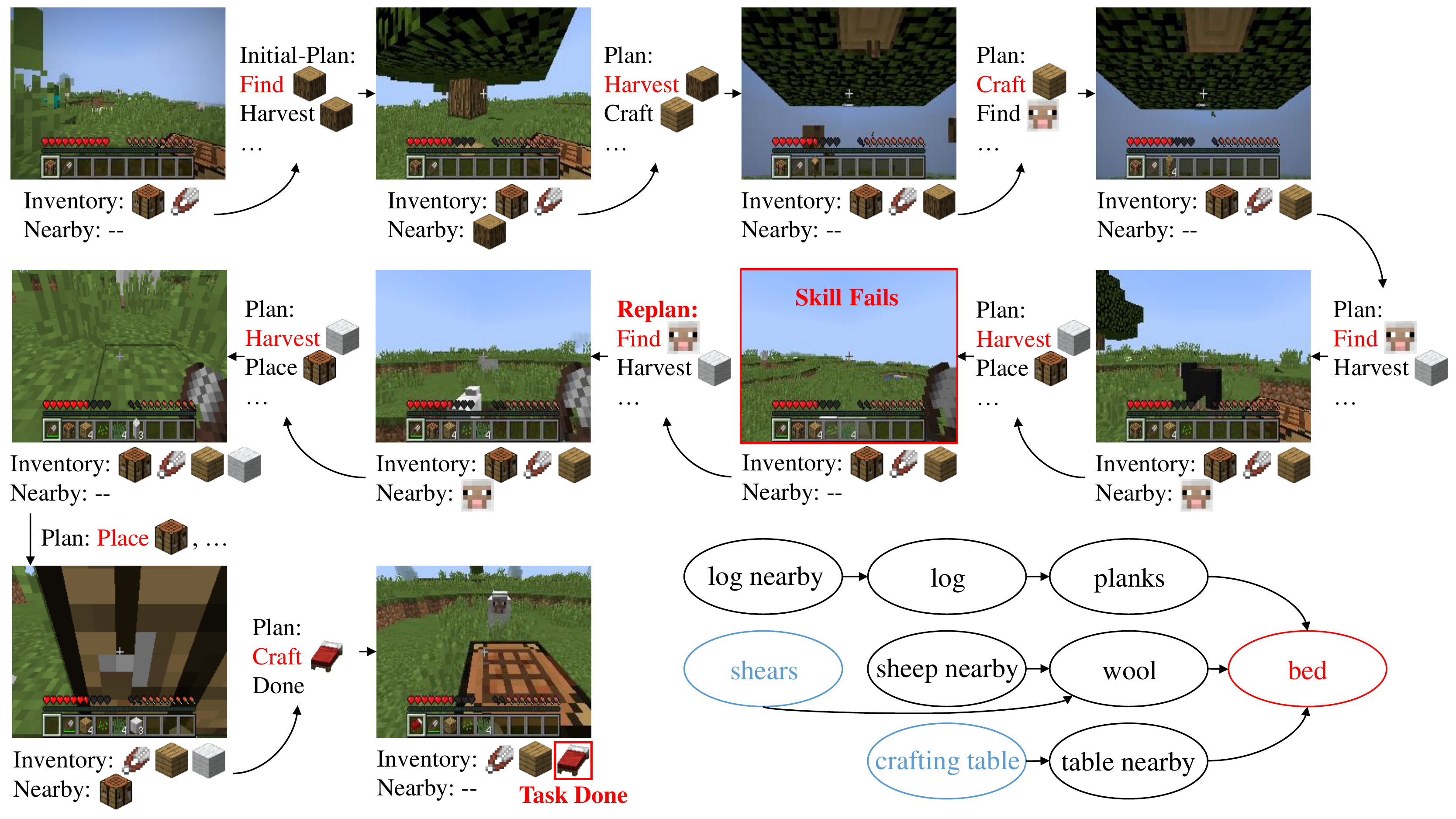}
  %\fbox{\rule[-.5cm]{0cm}{4cm} \rule[-.5cm]{4cm}{0cm}}
  \vspace{-4mm}
  \caption{Demonstration of a planning and execution episode for the task "craft a bed". Following the direction of the arrows, the planner iteratively proposes the skill sequence based on the agent's state, and the policy executes the first skill. Though an execution for "harvest wool" fails in the middle, the planner replans to "find a sheep" again to fix this error, and finally completes the task. The lower right shows the skill graph for this task, where the red circle indicates the target, and the blue circles indicate the initial items.}
  \label{fig:execute-pipeline}
\end{figure}

\begin{figure}[htbp]
  \centering
  \includegraphics[scale=0.4, trim={0cm, 9cm, 0cm, 0cm}, clip]{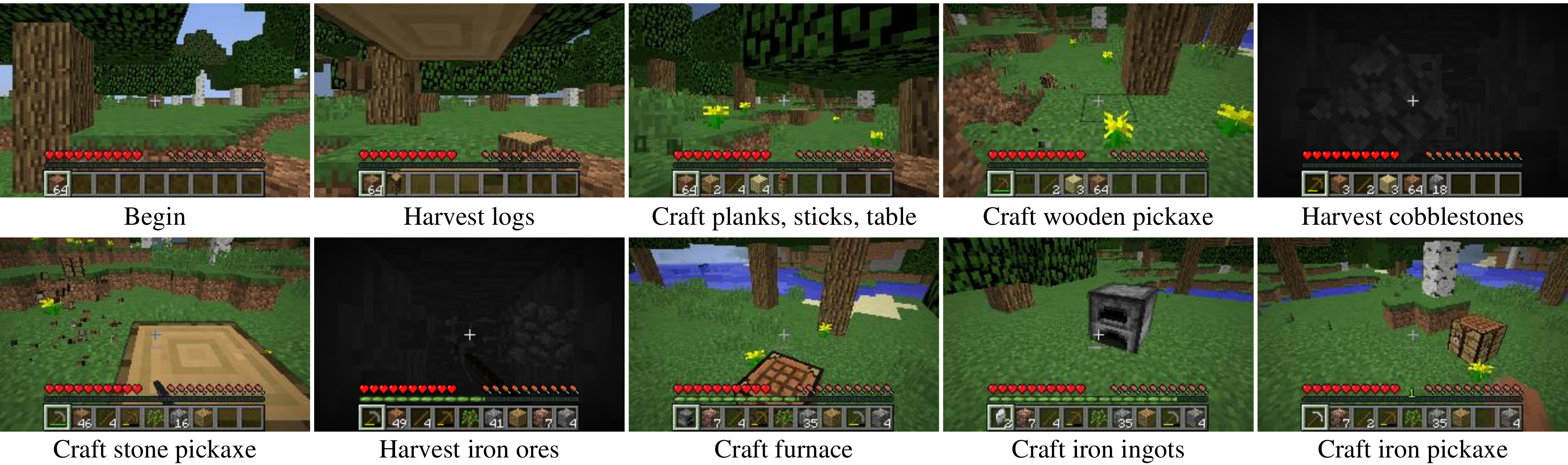}
  %\fbox{\rule[-.5cm]{0cm}{4cm} \rule[-.5cm]{4cm}{0cm}}
  \caption{A playing episode of Plan4MC for crafting iron pickaxe \textbf{with bare hands}. This is a challenging task in Minecraft Tech Tree, which requires 16 different basic skills and 117 steps in the initial plan.}
  \label{fig:ironpickaxe}
\end{figure}

Here we visually demonstrate the steps Plan4MC takes to solve a long-horizon task. Figure \ref{fig:execute-pipeline} shows the interactive planning and execution process for crafting a bed. Figure \ref{fig:ironpickaxe} shows the key frames of Plan4MC solving the challenging Tech Tree task of crafting an iron pickaxe with bare hands.

\vspace{2mm}

\newpage
\section{Algorithms}
\label{appendix:algorithms}

We present our algorithm sketches for skill planning and solving hard tasks here.

\begin{algorithm}[htbp]
	\caption{DFS.}
	\label{alg:dfs-sub}
	\KwIn{Pre-generated skill graph: $G$; Target item: $g$; Target item quantity: $n$;
 
    \textit{Global variables}: possessing items $I$ and skill sequence $S$.}
	\BlankLine
    \For{$g'$ in $parents(G,g)$}{
        $n_{g'}, n_{g}, consume \leftarrow <g',g>$;

        $n_{g'}^{todo}\leftarrow n_{g'}$;
        
        \If{(quantity of $g'$ in $I$) $> n_{g'}$}{
            Decrease $g'$ quantity with $n_{g'}$ in $I$, if $consume$;
        }
        \Else{
            $n_{g'}^{todo}\leftarrow n_{g'}^{todo} -$ (quantity of $g'$ in $I$);
        }
        \While{$n_{g'}^{todo}>0$}{
            DFS($G,g',n_{g'}^{todo}, I,S$);

            \If{$g'$ is not Crafting-skill}{
                Remove all nearby items in $I$;
            }
            $n_{g'}^{obtain}\leftarrow$ (quantity of $g'$ obtained after executing skill $g'$);
            
            \If{$n_{g'}^{obtain} > n_{g'}^{todo}$}{
                Increase $g'$ quantity with $n_{g'}^{obtain} - n_{g'}^{todo}$ in $I$;
            }
            Increase other obtained items after executing skill $g'$ in $I$;

            $n_{g'}^{todo} \leftarrow n_{g'}^{todo} - n_{g'}^{obtain}$;
        }
    }
    
	Append skill $g$ to $S$.
\end{algorithm}

\begin{algorithm}[htbp]
	\caption{Skill search algorithm.}
	\label{alg:dfs}
	\KwIn{Pre-generated skill graph: $G$; Target item: $g$; Initial items: $I$.}
	\KwOut{Skill sequence: $(s_1,s_2,...)$.}
	\BlankLine
    $S'\leftarrow ()$;
    
    $I'\leftarrow I$;

    DFS($G,g,1,I',S'$);
    
	return $S'$.
\end{algorithm}

\begin{algorithm}[htbp]
	\caption{Process for solving a task.}
	\label{alg:solve-task}
	\KwIn{Task: $T=(g,I)$; Pre-trained skills: $\{\pi_s\}_{s\in S}$; Pre-generated skill graph: $G$; Skill search algorithm: $Search$.}
	\KwOut{Task success.}
	\BlankLine
        $I'\leftarrow I$;
        
	\While{\textnormal{task not done}}{
		$(s_1,s_2,...) \leftarrow Search(G,g,I')$;
  
        Execute $\pi_{s_1}$ for several steps;

        \If{task success}{
            return \textbf{True};
        }

        $I'\leftarrow $ inventory items $\cup$ nearby items;
	}
	return \textbf{False}.
\end{algorithm}

\newpage

\section{Details in Training Basic Skills}
\label{appendix:skill-details}

Table \ref{tab:skill-train} shows the environment and algorithm configurations for training basic skills. Except for the skill of mining \mcironore\mcdiamond \  whose breaking speed multiplier in the simulator is set to 10, all the skills are trained using the unmodified MineDojo simulator. 

Though the MineCLIP reward improves the learning of many skills, it is still not enough to encourage some complicated behaviors. In combat \mccow \mcsheep, we introduce distance reward and attack reward to further encourage the agent to chase and attack the mobs. In mining \mclog \mccobblestone, we introduce distance reward to keep the agent close to the target blocks. To mine underground ores \mcironore \mcdiamond, we add depth reward to encourage the agent to mine deeper and then go back to the ground. These item-based intrinsic rewards are easy to implement for all the items and are also applicable in many other open-world environments like robotics.
Intrinsic rewards are implemented as follows.

\textbf{State count.} The high-level recurrent policy for Finding-skills optimizes the visited area in a $110\times 110$ square, where the agent's spawn location is at the center. We divide the square into $11\times 11$ grids and keep a visitation flag for each grid. Once the agent walks into an unvisited grid, it receives $+1$ state count reward.

\textbf{Goal navigation.} The low-level policy for Finding-skills is encouraged to reach the goal position. The goal location is randomly sampled in 4 directions at a distance of 10 from the agent. To get closer to the goal, we compute the distance change between the goal and the agent: $r_d = -(d_t-d_{t-1})$, where $d_t$ is the distance on the plane coordinates at time step $t$. Additionally, to encourage the agent to look in its walking direction, we add rewards to regularize the agent's yaw and pitch angles: $r_{yaw}=yaw\cdot g, r_{pitch}=cos(pitch)$, where $g$ is the goal direction. The total reward is: 
\begin{equation}
    r=r_{yaw}+r_{pitch}+10 * r_d.
\end{equation}

\textbf{CLIP reward.} This reward encourages the agent to produce behaviors that match the task prompt. We sample 31 task prompts among all the MineDojo programmatic tasks as negative samples. The pre-trained MineCLIP~\citep{minedojo} model computes the similarities between features of the past 16 frames and prompts. We compute the probability that the frames are most similar to the task prompt: $p=\left[\mathrm{softmax}\left(S\left(f_v, f_l\right), \{S\left(f_v,f_{l^-}\right)\}_{l^-}\right)\right]_0$, where $f_v, f_l$ are video features and prompt features, $l$ is the task prompt, and $l^-$ are negative prompts. The CLIP reward is:
\begin{equation}
    r_{\mathrm{CLIP}} = \max{\left\{p-\frac{1}{32}, 0\right\}}.
\end{equation}

\textbf{Distance.} The distance reward provides dense reward signals to reach the target items. For combat tasks, the agent gets a distance reward when the distance is closer than the minimal distance in history:
\begin{equation}
    r_{distance} = \max \left\{ \min_{t'<t}{d_{t'}} - d_t, 0 \right\}.
\end{equation}
For mining \mclog\mccobblestone \ tasks, since the agent should stay close to the block for many time steps, we modify the distance reward to encourage keeping a small distance:
\begin{equation}
r_{distance}=
\begin{cases}
        d_{t-1}-d_{t},  \quad 1.5\le d_t \le +\infty \\
        2, \quad d_t<1.5 \\
        -2, \quad d_t=+\infty,
\end{cases}
\end{equation}
where $d_t$ is the distance between the agent and the target item at time step $t$, which is detected by lidar rays in the simulator.

\textbf{Attack.} For combat tasks, we reward the agent for attacking the target mobs. We use the tool's durability information to detect valid attacks and use lidar rays to detect the target mob. The attack reward is:
\begin{equation}
r_{attack} = 
\begin{cases}
\begin{aligned}
 90, & \quad \text{if  valid attack and the target at center }  \\
 1, & \quad \text{if valid attack but the target not at center}  \\
 0, & \quad \text{otherwise. }
\end{aligned}
\end{cases}
\end{equation}

\textbf{Depth.} For mining \mcironore\mcdiamond tasks, the agent should dig down first, then go back to the ground. We use the y-axis to calculate the change of the agent's depth, and use the depth reward to encourage such behaviors. To train the dig-down policy, the depth reward is:
\begin{equation}
    r_{down} = \max \left\{ \min_{t'<t}{y_{t'}} - y_t, 0 \right\}.
\end{equation}
To train the go-back policy, the depth reward is:
\begin{equation}
    r_{up} = \max \left\{ y_t - \max_{t'<t}{y_{t'}}, 0 \right\}.
\end{equation}

For each Manipulation-skill, we use a linear combination of intrinsic reward and extrinsic success reward to train the policy.

\begin{table}[!t]
  \caption{Training configurations for all the basic skills. \textit{Max Steps} is the maximal episode length. \textit{Training Steps} shows the environment steps cost for training each skill.  \textit{Init.} shows the maximal distance to spawn mobs at environment reset. The high-level policy and low-level policy for Finding-skills are listed in two lines.}
  \label{tab:skill-train}
  \centering
  \begin{tabular}{ccccccc}
    \toprule
    Skill  & Max Steps & Method & Intrinsic Reward & Training Steps & Biome & Init. \\
    \midrule
    Find  &  \makecell{high: 40 \\ low: 50} & \makecell{PPO \\ DQN} & \makecell{state count \\ goal navigation} & \makecell{1M \\ 0.5M} & plains & \--\-- \\
    \midrule
    Place \mccraftingtable\mcfurnace & 200 & PPO & CLIP reward & 0.3M & \--\-- & \--\-- \\
    Harvest \mcmilkbucket & 200 & PPO & CLIP reward & 1M & plains & 10 \\
    Harvest \mcwool & 200 & PPO & CLIP reward & 1M & plains & 10 \\
    Combat \mccow & 400 & PPO & CLIP, distance, attack & 1M & plains & 2 \\
    Combat \mcsheep & 400 & PPO & CLIP, distance, attack & 1M & plains & 2 \\
    Harvest \mclog & 500 & PPO & distance & 0.5M & forest & \--\-- \\
    Harvest \mccobblestone & 1000 & PPO & distance & 0.3M & hills & \--\-- \\
    Mine \mcironore\mcdiamond & 50 & PPO & depth & 0.4M & forest & \--\-- \\
    \midrule
    Craft & 1 & \--\-- & \--\-- & 0 & \--\-- & \--\-- \\
    \bottomrule
  \end{tabular}
\end{table}

\begin{table}[!t]
  \caption{Information for all the selected basic skill policies. \textit{Success Rate} is the success rate of the selected policy on the smoothed training curve.}
  \label{tab:skill-info}
  \centering
  \begin{tabular}{cccc}
    \toprule
    Skill  & Parameters & Execute Steps & Success Rate \\
    \midrule
    Find  & 0.9M & 1000 & \--\-- \\
    \midrule
    Place \mccraftingtable\mcfurnace & 2.0M & 200 & 0.98 \\
    Harvest \mcmilkbucket & 2.0M & 200 & 0.50 \\
    Harvest \mcwool & 2.0M & 200 & 0.27 \\
    Combat \mccow & 2.0M & 400 & 0.21 \\
    Combat \mcsheep & 2.0M & 400 & 0.30 \\
    Harvest \mclog & 2.0M & 500 & 0.56 \\
    Harvest \mccobblestone & 2.0M & 200 & 0.47 \\
    Mine \mcironore\mcdiamond & 4.0M & 1000 & -- \\
    \midrule
    Craft & 0 & 1 & 1.00  \\
    \bottomrule
  \end{tabular}
\end{table}

It takes one day on a single TITAN Xp GPU to train each skill for 1M environmental steps.
Table \ref{tab:skill-info} shows our selected basic skill policies for downstream tasks. Since the Finding-skill and the Mining \mcironore\mcdiamond \ skill has no success rate during training, we pick the models with the highest returns on the smoothed training curves. For other skills, we pick the models with the highest success rates on the smoothed training curves.

\section{LLM Prompt Design}
\label{appendix:llmprompt}

To construct the skill graph, we use ChatGPT to generate skill information showing relationships between skills. For both Manipulation-skills and Crafting-skills, we demonstrate several skills, explain the format, and then ask ChatGPT to generate all the skills in the same format. 

\begin{lstlisting}[caption=Prompt for generating some Manipulation-skills and ChatGPT's response. We only modify the last querying sentence in the prompt to generate all the skills. ]
`\underline{\small\textbf{Prompt:}}` 
I am playing the game Minecraft. I define some basic skills, like attack something, collect something and place something nearby. I list the skills in a special format. 
As an example:
furnace_nearby:
  consume: 
    'furnace': 1
  require: 
  equip: ['furnace']
  obtain: 
    'furnace_nearby': 1
To understand this skill line by line: the skill is to get a furnace_nearby. 'consume' means things will be consumed or killed. In this skill, furnace*1 will be consumed.  'require' means things are needed but will not be consumed. In this skill, nothing else is required. We should equip furnace to the first slot. If you do not have to equip anything, write 'equip: []'. Finally, we will obtain furnace_nearby*1. 
Another example:
cobblestone:
  consume: 
    'cobblestone_nearby': 1
  require: 
    'wooden_pickaxe': 1
  equip: ['wooden_pickaxe']
  obtain: 
    'cobblestone': 1
To understand: to mine a cobblestone, we will consume a nearby cobblestone. A wooden_pickaxe is required and should be equipped, but will not be consumed.
Now you understand the rule of this format. Please help me generate the following skills:
crafting_table_nearby, wool, beef, diamond

`\underline{\small\textbf{ChatGPT:}}`
Here are the generated skills in the specified format:

crafting_table_nearby:
consume:
'crafting_table': 1
require:
equip: ['crafting_table']
obtain:
'crafting_table_nearby': 1

wool:
consume:
'sheep_nearby': 1
require:
'shears': 1
equip: ['shears']
obtain:
'wool': 1

beef:
consume:
'cow_nearby': 1
require:
equip: []
obtain:
'beef': 1

diamond:
consume:
'diamond_ore_nearby': 1
require:
'iron_pickaxe': 1
equip: ['iron_pickaxe']
obtain:
'diamond': 1
\end{lstlisting}

\begin{lstlisting}[caption=Prompt for generating some Crafting-skills and ChatGPT's response. We only modify the last querying sentence in the prompt to generate all the skills.]
`\underline{\small\textbf{Prompt:}}` 
I am playing the game Minecraft. I list all the crafting skills in this game in a special format. 
As an example:
bed:
  consume: 
    'planks': 3 
    'wool': 3
  require: 
    'crafting_table_nearby': 1
  equip: []
  obtain: 
    'bed': 1
To understand this skill line by line: the skill is to craft a bed. We will consume planks*3 and wool*3. To craft bed, a crafting_table_nearby is necessary and must be in the 'require:' line.
For crafting skills the equipment is always []. Finally, we will obtain bed*1. 
Another example:
cooked_beef:
  consume:
    'planks': 1
    'beef': 1
  require:
    'furnace_nearby': 1
  equip: []
  obtain:
    'cooked_beef': 1
Because we require a furnace_nearby. The fuel, which is always 'planks'*1, should be added to the 'consume' items.
For the 'require:' line, you should know whether the skill needs crafting_table_nearby,  furnace_nearby, or nothing. If require nothing, just don't write anything in this line.
Now you understand the rule of this format. Please help me generate the following skills:
stick, chest, furnace,item_frame

`\underline{\small\textbf{ChatGPT:}}`
Sure, here are the crafting skills for the items you requested:

stick:
consume:
'planks': 2
require:
equip: []
obtain:
'stick': 4

chest:
consume:
'planks': 8
require:
'crafting_table_nearby': 1
equip: []
obtain:
'chest': 1

furnace:
consume:
'cobblestone': 8
require:
'crafting_table_nearby': 1
equip: []
obtain:
'furnace': 1

item_frame:
consume:
'stick': 8
'leather': 1
require:
'crafting_table_nearby': 1
equip: []
obtain:
'item_frame': 1
\end{lstlisting}

All the Finding-skills have the same format:
\begin{lstlisting}
(item)_nearby:
    consume:
    require:
    equip: []
    obtain:
        (item)_nearby: 1
\end{lstlisting}

In our experiments, after generating all the skills, we check and correct them according to the Minecraft database. Among 55 skills, we find 6 mistakes in total.

\section{Task Setup}
\label{appendix:task-setup}

Table \ref{tab:task-setup}, \ref{tab:task-setup-2} lists settings for 40 evaluation tasks. To make sure the agent is spawned in an unseen environment in each test episode, we randomly transport the agent with a maximum distance of 500 at environment reset. For tasks involving interacting with mobs, we spawn cows and sheep with a maximum distance of 30, which is much larger than the spawning distance in training basic skills. For the Mine-Ores task set, we set the breaking speed multiplier to 10. For the other three task sets, we use the default simulator.

\begin{table}[htbp]
  \caption{Settings for Cut-Trees and Mine-Stones tasks. \textit{Initial Tools} are provided in the inventory at each episode beginning. \textit{Involved Skills} is the least number of basic skills the agent should master to accomplish the task. \textit{Planning Steps} is the number of basic skills to be executed sequentially in the initial plans. }
  \label{tab:task-setup}
  \centering
  \begin{tabular}{ccccccc}
    \toprule
    Task Icon  & Target Name & Initial Tools & Biome & Max Steps & \makecell{Involved\\Skills} & \makecell{Planning\\Steps} \\
    \midrule
    \mcstick  & stick & \--\-- & plains & 3000 & 4 & 4  \\
    \mccraftingtable &  \makecell{crafting\_table\_ \\ nearby} & \--\-- & plains & 3000 & 5 & 5 \\
    \mcbowl  &  bowl & \--\-- & forest & 3000 & 6 & 9 \\
    \mcchest   & chest  & \--\-- & forest & 3000 & 6 & 12  \\
    \mctrapdoor  & trap\_door  & \--\-- & forest & 3000 & 6 & 12  \\
    \mcsign  &  sign & \--\-- & forest & 3000 & 7 & 13  \\
    \mcwoodenshovel  &  wooden\_shovel & \--\-- & forest & 3000 & 7 & 10 \\
    \mcwoodensword  &  wooden\_sword & \--\-- & forest & 3000 & 7 & 10  \\
    \mcwoodenaxe  &  wooden\_axe & \--\-- & forest & 3000 & 7 & 13  \\
    \mcwoodenpickaxe  &  wooden\_pickaxe & \--\-- & forest & 3000 & 7 & 13  \\
    \midrule
    \mcfurnace  &  furnace\_nearby & \mclog*10 & hills & 5000 & 9 & 28 \\
    \mcstonestairs  & stone\_stairs  & \mclog*10 & hills & 5000 & 8 & 23 \\
    \mcstoneslab  & stone\_slab  &  \mclog*10 & hills & 3000 & 8 & 17  \\
    \mccobblestonewall  & cobblestone\_wall  & \mclog*10 & hills & 5000 & 8 & 23  \\
    \mclever  &  lever & \mcwoodenpickaxe & forest\_hills & 5000 & 7 & 7 \\
    \mctorch  & torch  & \mclog*10 & hills & 5000 & 11 & 30  \\
    \mcstoneshovel  &  stone\_shovel & \mcwoodenpickaxe & forest\_hills & 10000 & 9 & 12 \\
    \mcstonesword  &  stone\_sword & \mcwoodenpickaxe & forest\_hills & 10000 & 9 & 14 \\
    \mcstoneaxe  &  stone\_axe & \mcwoodenpickaxe & forest\_hills & 10000 & 9 & 16 \\
    \mcstonepickaxe  &  stone\_pickaxe & \mcwoodenpickaxe & forest\_hills & 10000 & 9 & 16 \\
    \bottomrule
  \end{tabular}
\end{table}

\begin{table}[htbp]
  \caption{Settings for Mine-Ores and Interact-Mobs tasks. \textit{Initial Tools} are provided in the inventory at each episode beginning. \textit{Involved Skills} is the least number of basic skills the agent should master to accomplish the task. \textit{Planning Steps} is the number of basic skills to be executed sequentially in the initial plans. }
  \label{tab:task-setup-2}
  \centering
  \begin{tabular}{ccccccc}
    \toprule
    Task Icon  & Target Name & Initial Tools & Biome & Max Steps & \makecell{Involved\\Skills} & \makecell{Planning\\Steps} \\
    \midrule
    \mcironingot  & iron\_ingot & \mcstonepickaxe*5, \mcdirt*64 & forest & 8000 & 12 & 30  \\
    \mctripwirehook  & tripwire\_hook & \mcstonepickaxe*5, \mcdirt*64 & forest & 8000 & 14 & 35  \\
    \mcheavypressureplate  & \makecell{heavy\_weighted\_ \\ pressure\_plate} & \mcstonepickaxe*5, \mcdirt*64 & forest & 10000 & 13 & 61  \\
    \mcshears  & shears & \mcstonepickaxe*5, \mcdirt*64 & forest & 10000 & 13 & 61  \\
    \mcbucket  & bucket & \mcstonepickaxe*5, \mcdirt*64 & forest & 12000 & 13 & 91  \\
    \mcirontrapdoor  & iron\_trapdoor & \mcstonepickaxe*5, \mcdirt*64 & forest & 12000 & 13 & 121  \\
    \mcironshovel  & iron\_shovel & \mcstonepickaxe*5, \mcdirt*64 & forest & 8000 & 14 & 35  \\
    \mcironsword  & iron\_sword & \mcstonepickaxe*5, \mcdirt*64 & forest & 10000 & 14 & 65  \\
    \mcironaxe  & iron\_axe & \mcstonepickaxe*5, \mcdirt*64 & forest & 12000 & 14 & 95  \\
    \mcironpickaxe  & iron\_pickaxe & \mcstonepickaxe*5, \mcdirt*64 & forest & 12000 & 14 & 95  \\
    \midrule
    \mcmilkbucket  & milk\_bucket & \mccraftingtable, \mcironingot*3 & plains & 3000 & 4 & 4 \\
    \mcwool  & wool & \mccraftingtable, \mcironingot*2 & plains & 3000 & 3 & 3 \\
    \mcbeef  & beef  & \mcdiamondsword & plains & 3000 & 2 & 2 \\
    \mcmutton  & mutton & \mcdiamondsword & plains & 3000 & 2 & 2 \\
    \mcbed  &  bed & \mccraftingtable, \mcshears & plains & 10000 & 7 & 11 \\
    \mcpainting  & painting & \mccraftingtable, \mcshears & plains & 10000 & 8 & 9 \\
    \mccarpet  & carpet  & \mcshears & plains & 3000 & 3 & 5 \\
    \mcitemframe  & item\_frame & \mccraftingtable, \mcdiamondsword & plains & 10000 & 8 & 9 \\
    \mccookedbeef  & cooked\_beef & \mcfurnace, \mcdiamondsword & plains & 10000 & 7 & 7 \\
    \mccookedmutton  & cooked\_mutton & \mcfurnace, \mcdiamondsword & plains & 10000 & 7 & 7 \\
    \bottomrule
  \end{tabular}
\end{table}

\section{Experiment Results for All the Tasks}
Table \ref{tab:results-all-tasks} shows the success rates of all the methods in all the tasks, grouped in 4 task sets.

\label{appendix:results-all}
\begin{table}[htbp]
  \caption{Success rates in all the tasks. Each task is tested for 30 episodes, set with the same random seeds across different methods.}
  \label{tab:results-all-tasks}
  \centering
  \begin{tabular}{ccccccc}
    \toprule
    Task  & MineAgent & \makecell{Plan4MC w/o\\ Find-skill} & \makecell{Interactive\\ LLM} & \makecell{Plan4MC\\ Zero-shot} & \makecell{Plan4MC\\ 1/2-steps} & Plan4MC \\
    \midrule
    \mcstick  &  0.00 & 0.03 & 0.30 & 0.27 & 0.30 & 0.30  \\
    \mccraftingtable   & 0.03 & 0.07 & 0.17 & 0.27 & 0.20 & 0.30 \\
    \mcbowl  &  0.00 & 0.40 & 0.07 & 0.27 & 0.57 & 0.47 \\
    \mcchest   & 0.00  & 0.23 & 0.00 & 0.07 & 0.10 & 0.23 \\
    \mctrapdoor  & 0.00  & 0.07 & 0.03 & 0.20 & 0.27 & 0.37 \\
    \mcsign  &  0.00 & 0.07 & 0.00 & 0.10 & 0.30 & 0.43 \\
    \mcwoodenshovel & 0.00 & 0.37 & 0.73 & 0.23 & 0.50 & 0.70 \\
    \mcwoodensword & 0.00 & 0.33 & 0.63 & 0.30 & 0.60 & 0.47 \\
    \mcwoodenaxe & 0.00 & 0.23 & 0.47 & 0.13 & 0.27 & 0.37 \\
    \mcwoodenpickaxe  & 0.00 & 0.07 & 0.20 & 0.00 & 0.27 & 0.53 \\
    \midrule
    \mcfurnace  &  0.00 & 0.17 & 0.00 & 0.00 & 0.13 & 0.37 \\
    \mcstonestairs  & 0.00  & 0.30 & 0.20 & 0.00 & 0.33 & 0.47 \\
    \mcstoneslab  & 0.00  & 0.20 & 0.03 & 0.00 & 0.37 & 0.53 \\
    \mccobblestonewall  & 0.21 & 0.13 & 0.13 & 0.00 & 0.33 & 0.57 \\
    \mclever  &  0.00 & 0.00 & 0.00 & 0.00 & 0.10 & 0.10 \\
    \mctorch  & 0.05 & 0.10 & 0.00 & 0.00 & 0.17 & 0.37 \\
    \mcstoneshovel & 0.00 & 0.00 & 0.10 & 0.00 & 0.03 & 0.20 \\
    \mcstonesword & 0.00 & 0.07 & 0.13 & 0.00 & 0.07 & 0.10 \\
    \mcstoneaxe & 0.00 & 0.00 & 0.07 & 0.00 & 0.10 & 0.07 \\
    \mcstonepickaxe  & 0.00  & 0.00 & 0.00 & 0.00 & 0.00 & 0.17 \\  
    \midrule
    \mcironingot & 0.00 & 0.53 & 0.20 & 0.00 & 0.30 & 0.47 \\
    \mctripwirehook & 0.00 & 0.27 & 0.00 & 0.00 & 0.27 & 0.33 \\
    \mcheavypressureplate & 0.00 & 0.37 & 0.00 & 0.00 & 0.13 & 0.30 \\
    \mcshears & 0.00 & 0.30 & 0.03 & 0.00 & 0.20 & 0.43 \\
    \mcbucket & 0.00 & 0.27 & 0.00 & 0.00 & 0.03 & 0.20 \\
    \mcirontrapdoor & 0.00 & 0.10 & 0.00 & 0.00 & 0.03 & 0.13  \\
    \mcironshovel & 0.00 & 0.27 & 0.03 & 0.00 & 0.27 & 0.37 \\
    \mcironsword & 0.00 & 0.13 & 0.00 & 0.00 & 0.07 & 0.20 \\
    \mcironaxe & 0.00 & 0.07 & 0.03 & 0.00 & 0.07 & 0.07 \\
    \mcironpickaxe & 0.00 & 0.13 & 0.00 & 0.00 & 0.07 & 0.17 \\
    \midrule
    \mcmilkbucket  &  0.46 & 0.57 & 0.57 & 0.60 & 0.63 & 0.83 \\
    \mcwool  & 0.50  & 0.40 & 0.76 & 0.30 & 0.60 & 0.53 \\
    \mcbeef  &  0.33 & 0.23 & 0.43 & 0.10 & 0.27 & 0.43 \\
    \mcmutton  &  0.35 & 0.17 & 0.30 & 0.07 & 0.13 & 0.33 \\
    \mcbed  & 0.00  & 0.00 & 0.00 & 0.00 & 0.07 & 0.17 \\
    \mcpainting  & 0.00  & 0.03 & 0.00 & 0.10 & 0.23 & 0.13 \\
    \mccarpet  & 0.06  & 0.27 & 0.37 & 0.10 & 0.50 & 0.37 \\
    \mcitemframe  & 0.00  & 0.00 & 0.00 & 0.03 & 0.10 & 0.07 \\
    \mccookedbeef  &  0.00 & 0.03 & 0.03 & 0.03 & 0.20 & 0.20 \\
    \mccookedmutton  & 0.00  & 0.00 & 0.00 & 0.00 & 0.03 & 0.13 \\
    \bottomrule
  \end{tabular}
\end{table}

\section{Training Manipulation-skills without Nearby Items}
\label{appendix:go-explore}

For all the Manipulation-skills that are trained with specified environments in the paper, we use the Go-Explore-like approach to re-train them in the environments without target items initialized nearby. 
In a training episode, the pre-trained Finding-skill explores the environment and finds the target item, then the policy collects data for RL training. In the following, we denote the previous method as Plan4MC and the new method as Plan4MC-go-explore.

Table \ref{tab:skill-success-go-explore} shows the maximal success rates of these skills over 100 training epochs. We find that all the skills trained with Go-Explore do not fail and the success rates are comparable to the previous skills. This is because the Finding-skill provides good environmental initialization for the training policies. In Milk and Wool, Plan4MC-go-explore even outperforms Plan4MC, because the agent can be closer to the target mobs in Plan4MC-go-explore.

Table \ref{tab:task-success-go-explore} shows the test performance of Plan4MC on the four task sets. We find that Plan4MC-go-explore even outperforms Plan4MC on three task sets. This demonstrates that the skills trained with Go-Explore can generalize well to unseen environments. %We argue that the Find-skill randomly finds different target individuals, approaches the target from different angles, and the mobs move in the environment, enabling training in sufficiently diverse scenes and generalize.

\begin{table}[htbp]
  \caption{Training success rates of the Manipulation-skills under the two environment settings. Results are the maximal success rates averaged on 100 training epochs.}
  \label{tab:skill-success-go-explore}
  \centering
  \begin{tabular}{ccccccc}
    \toprule
    %\multirow{2}{*}{Skill}  &	Milk &	Wool &	Cow &	Sheep &	Log &	Cobblestone \\
    Skill             & \mcmilkbucket & \mcwool & \mcbeef & \mcmutton & \mclog & \mccobblestone \\
    \midrule
    Plan4MC &	0.50 &	0.27 &	0.21 &	0.30 &	0.56 &	0.47 \\
    Plan4MC-go-explore &	0.82 &	0.34 &	0.22 &	0.19 &	0.25 &	0.71 \\
    \bottomrule
  \end{tabular}
\end{table}

\begin{table}[htbp]
  \caption{Average success rates on the four task sets of Plan4MC, with the Manipulation-skills trained in the two settings.}
  \label{tab:task-success-go-explore}
  \centering
  \begin{tabular}{ccccc}
    \toprule
    Task Set &	Cut-Trees &	Mine-Stones &	Mine-Ores &	Interact-Mobs \\
    \midrule
    Plan4MC &	0.417 &	0.293 &	0.267 &	0.320 \\
    Plan4MC-go-explore &	0.543 &	0.349 &	0.197 &	0.383 \\
    \bottomrule
  \end{tabular}
\end{table}

We further study the generalization capabilities of learned skills. Table \ref{tab:skill-generalization-go-explore} shows the test success rates of these skills in the 40 tasks and the generalization gap. %As we explained in the paper, this gap can be positive because subtasks may be easier in test than in training.
We observe that Plan4MC-go-explore has a small generalization gap in the first four mob-related skills. This is because Plan4MC-go-explore uses the same policy for approaching the target mob in training and test, yielding closer initial distributions for Manipulation-skills. We find that in Harvest Log, Plan4MC-go-explore often finds trees that have been cut before. Thus, it is more difficult to harvest logs in training, and the test success rate exceeds the training success rate.

\begin{table}[htbp]
  \caption{The test success rates of the skills in solving the 40 tasks, and the generalization gap (test success rate - training success rate).}
  \label{tab:skill-generalization-go-explore}
  \centering
  \begin{tabular}{ccccccc}
    \toprule
    %\multirow{2}{*}{Skill}  &	Milk &	Wool &	Cow &	Sheep &	Log &	Cobblestone \\
     Skill            & \mcmilkbucket & \mcwool & \mcbeef & \mcmutton & \mclog & \mccobblestone \\
    \midrule
    Plan4MC &	0.71(+0.21) &	0.26(-0.01) &	0.27(+0.06) &	0.16(-0.14) &	0.33(-0.23) &	0.26(-0.21) \\
    \makecell{Plan4MC- \\ go-explore} &	0.86(-0.04) &	0.47(+0.13) &	0.16(-0.06) &	0.16(-0.03) &	0.45(+0.20) &	0.47(-0.24) \\
    \bottomrule
  \end{tabular}
\end{table}

\section{Discussion on the Generalization of Plan4MC}
\label{appendix:discussion}

Plan4MC contributes a pipeline combining LLM-assisted planning and RL for skill acquisition. It is widely applicable in many open-world domains~\citep{saycan, behavior1k}, where the agent can combine basic skills to solve diverse long-horizon tasks.

Our key insight is that we can divide a skill into fine-grained basic skills, thus enabling acquiring skills sample-efficiently with demonstration-free RL. The Finding-skill in Plan4MC can be replaced with any learning-to-explore RL policy, or a navigation policy in robotics.
As an example, for indoor robotic tasks, a skill is defined with action (pick/drop/open) + object. We can break such a skill into navigation, arm positioning, and object manipulation, which can be better acquired with demonstration-free RL since the exploration difficulty is substantially reduced.

\begin{figure}[htbp]
  \centering
  \includegraphics[scale=0.14, trim={0cm, 0cm, 0cm, 0cm}, clip]{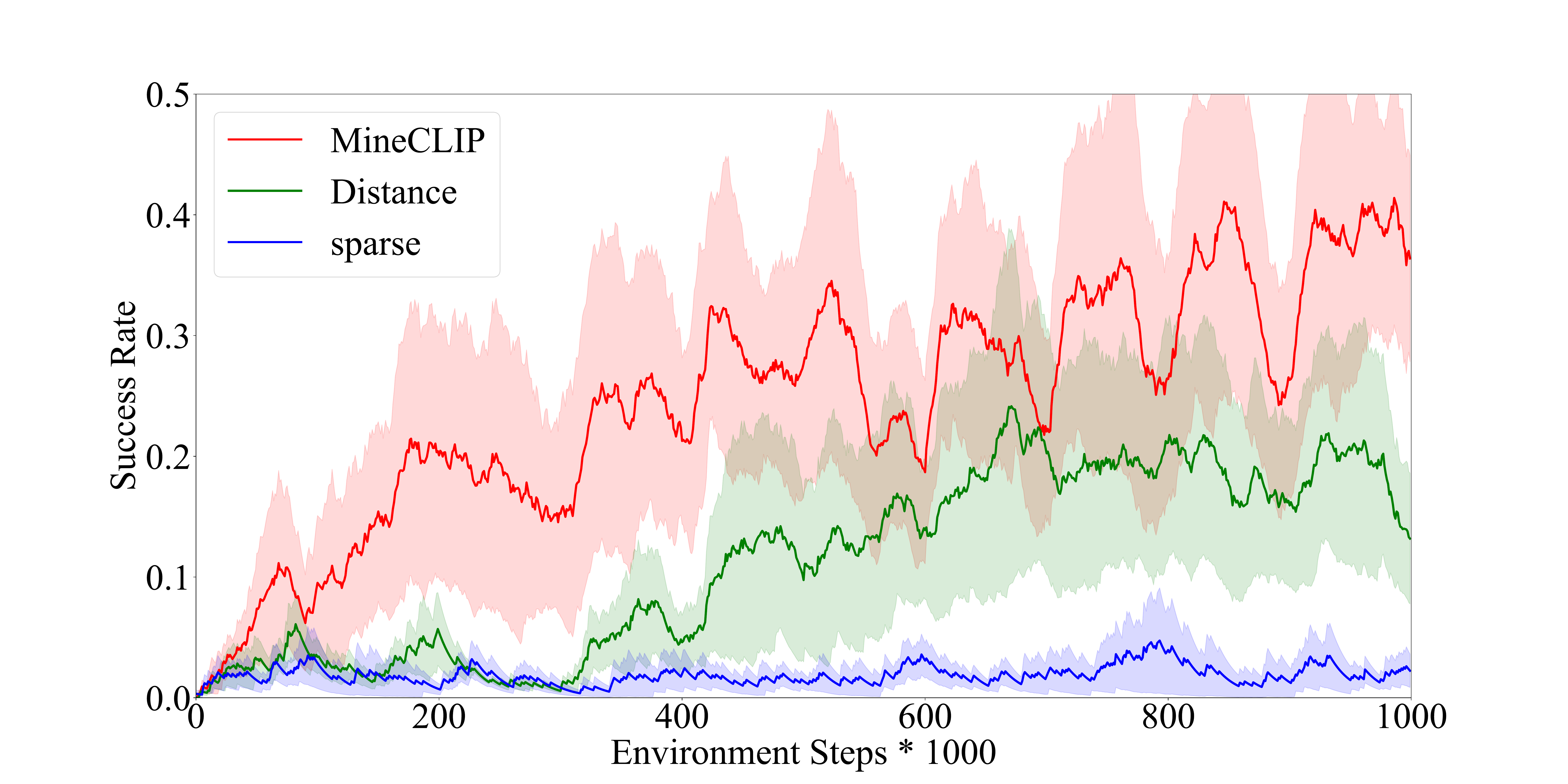}
  %\fbox{\rule[-.5cm]{0cm}{4cm} \rule[-.5cm]{4cm}{0cm}}
  \vspace{-2mm}
  \caption{Using different intrinsic rewards for training Harvest Milk with PPO. Results are averaged on 3 random seeds.}
  \label{fig:intrinsic-rewards}
\end{figure}

Our experiments on learning skills in Minecraft demonstrate that object-based intrinsic rewards improve sample efficiency. Figure \ref{fig:intrinsic-rewards} shows that both MineCLIP reward and distance reward have a positive impact on skill reinforcement learning. This gives motivation to use vision-language models, object detectors, or distance estimation for reward design in skill learning.

For planning, our approach is a novel extension of LLM-based planners, which incorporates LLM knowledge into a graph-based planner, improving planning accuracy. It can be extended to settings where the agent's state can be abstracted by text or entities.

\end{document}